\documentclass{article}
\usepackage{ijcai19}

\usepackage{times}
\usepackage{soul}
\usepackage{url}
\usepackage[hidelinks]{hyperref}
\usepackage[utf8]{inputenc}
\usepackage[small]{caption}
\usepackage{graphicx}
\usepackage{amsmath}
\usepackage{booktabs}
\usepackage{algorithm}
\usepackage{algorithmic}
\title{A Survey on Neural Network Language Models}

\author{
Kun Jing\footnote{K. Jing and J. Xu are corresponding authors.}
\And
Jungang Xu$^*$
\affiliations
School of Computer Science and Technology, University of Chinese Academy of Sciences, Beijing
\emails
jingkun18@mails.ucas.ac.cn,
xujg@ucas.ac.cn
}

\begin{document}

\maketitle

\begin{abstract}
  As the core component of Natural Language Processing (NLP) system, Language Model (LM) can provide word representation and probability indication of word sequences. Neural Network Language Models (NNLMs) overcome the curse of dimensionality and improve the performance of traditional LMs. A survey on NNLMs is performed in this paper. The structure of classic NNLMs is described firstly, and then some major improvements are introduced and analyzed. We summarize and compare corpora and toolkits of NNLMs. Furthermore, some research directions of NNLMs are discussed.
\end{abstract}

\section{Introduction}
LM is the basis of various NLP tasks. For example, in Machine Translation tasks, LM is used to evaluate probabilities of outputs of the system to improve fluency of the translation in the target language. In Speech Recognition tasks, LM and acoustic model are combined to predict the next word.

Early NLP systems were based primarily on manually written rules, which is time-consuming and laborious and cannot cover a variety of linguistic phenomena. In the 1980s, statistical LMs were proposed, which assigns probabilities to a sequence $s$ of $N$ words, i.e.,
\begin{align}
    P(s) =& P(w_1w_2\cdots w_N) \nonumber\\ =& P(w_1)P(w_2|w_1)\cdots P(w_N|w_1w_2\cdots w_{N-1}),
    \label{formula:LMgoal}
\end{align}
where $w_i$ denotes $i$-th word in the sequence $s$. The probability of a word sequence can be broken into the product of the conditional probability of the next word given its predecessors that are generally called a history of context or context.

Considering that it is difficult to learn the extremely many parameters of the above model, an approximate method is necessary. N-gram model is an approximation method, which was the most widely used and the state-of-the-art model before NNLMs. A (k+1)-gram model is derived from the k-order Markov assumption. This assumption illustrates that the current state depends only on the previous $k$ states, i.e.,
\begin{align}
    P(w_t|w_1\cdots w_{t-1}) \approx P(w_t|w_{t-k}\cdots w_{t-1}),
    \label{formula:approx}
\end{align}
which are estimated by maximum likelihood estimation.

Perplexity (PPL) \cite{jelinek_perplexitymeasure_1977}, an information-theoretic metric that measures the quality of a probabilistic model, is a way to evaluate LMs. Lower PPL indicates a better model. Given a corpus containing $N$ words and a language model $LM$, the PPL of $LM$ is
\begin{align}
    2^{-\frac{1}{N}\sum_{t=1}^Nlog_2LM(w_t|w_1\cdots w_{t-1})}.
\end{align}
It is noted that PPL is related to corpus. Two or more LMs can be compared on the same corpus in terms of PPL.

However, n-gram LMs have a significant drawback. The model would assign probabilities of 0 to the n-grams that do not appear in the training corpus, which does not match the actual situation. Smoothing techniques can solve this problem. Its main idea is {\em robbing the rich for the poor}, i.e., reducing the probability of events that appear in the training corpus and assigning the probability to events that do not appear.

Although n-gram LMs with smoothing techniques work out, there still are other problems. A fundamental problem is the curse of dimensionality, which limits modeling on larger corpora for a universal language model. It is particularly obvious in the case when one wants to model the joint distribution in discrete space. For example, if one wants to model an n-gram LM with a vocabulary of size $10,000$, there are potentially $10000^n-1$ free parameters.

In order to solve this problem, Neural Network (NN) is introduced for language modeling in continuous space. NNs including Feedforward Neural Network (FFNN), Recurrent Neural Network (RNN), can automatically learn features and continuous representation. Therefore, NNs are expected to be applied to LMs, even other NLP tasks, to cover the discreteness, combination, and sparsity of natural language.

The first FFNN Language Model (FFNNLM) presented by \cite{bengio_neural_2003} fights the curse of dimensionality by learning a distributed representation for words, which represents a word as a low dimensional vector, called embedding. FFNNLM performs better than n-gram LM. Then, RNN Language Model (RNNLM) \cite{mikolov_recurrent_2010} also was proposed. Since then, the NNLM has gradually become the mainstream LM and has rapidly developed. Long Short-term Memory RNN Language Model (LSTM-RNNLM) \cite{sundermeyer_lstm_2012} was proposed for the difficulty of learning long-term dependence. Various improvements were proposed for reducing the cost of training and evaluation and PPL such as hierarchical Softmax, caching, and so on. Recently, attention mechanisms have been introduced to improve NNLMs, which achieved significant performance improvements.

In this paper, we concentrate on reviewing the methods and trends of NNLM. Classic NNLMs are described in Section 2. Different types of improvements are introduced and analyzed separately in Section 3. Corpora and toolkits are described in Section 4 and Section 5. Finally, conclusions are given, and new research directions of NNLMs are discussed.
\section{Classic Neural Network Language Models}
\subsection{FFNN Language Models}
\cite{xu_can_2000} tried to introduce NNs into LMs. Although their model performs better than the baseline n-gram LM, their model with poor generalization ability cannot capture context-dependent features due to no hidden layer.

According to Formula \ref{formula:LMgoal}, the goal of LMs is equivalent to an evaluation of the conditional probability $P(w_k|w_1\cdots w_{k-1})$. But the FFNNs cannot directly process variable-length data and effectively represent the historical context. Therefore, for sequence modeling tasks like LMs, FFNNs have to use fixed-length inputs. Inspired by the n-gram LMs (see Formula \ref{formula:approx}), FFNNLMs consider the previous $n-1$ words as the context for predicting the next word.

\begin{figure}
	\centering
	\includegraphics[width=8cm,trim=0 200 0 200]{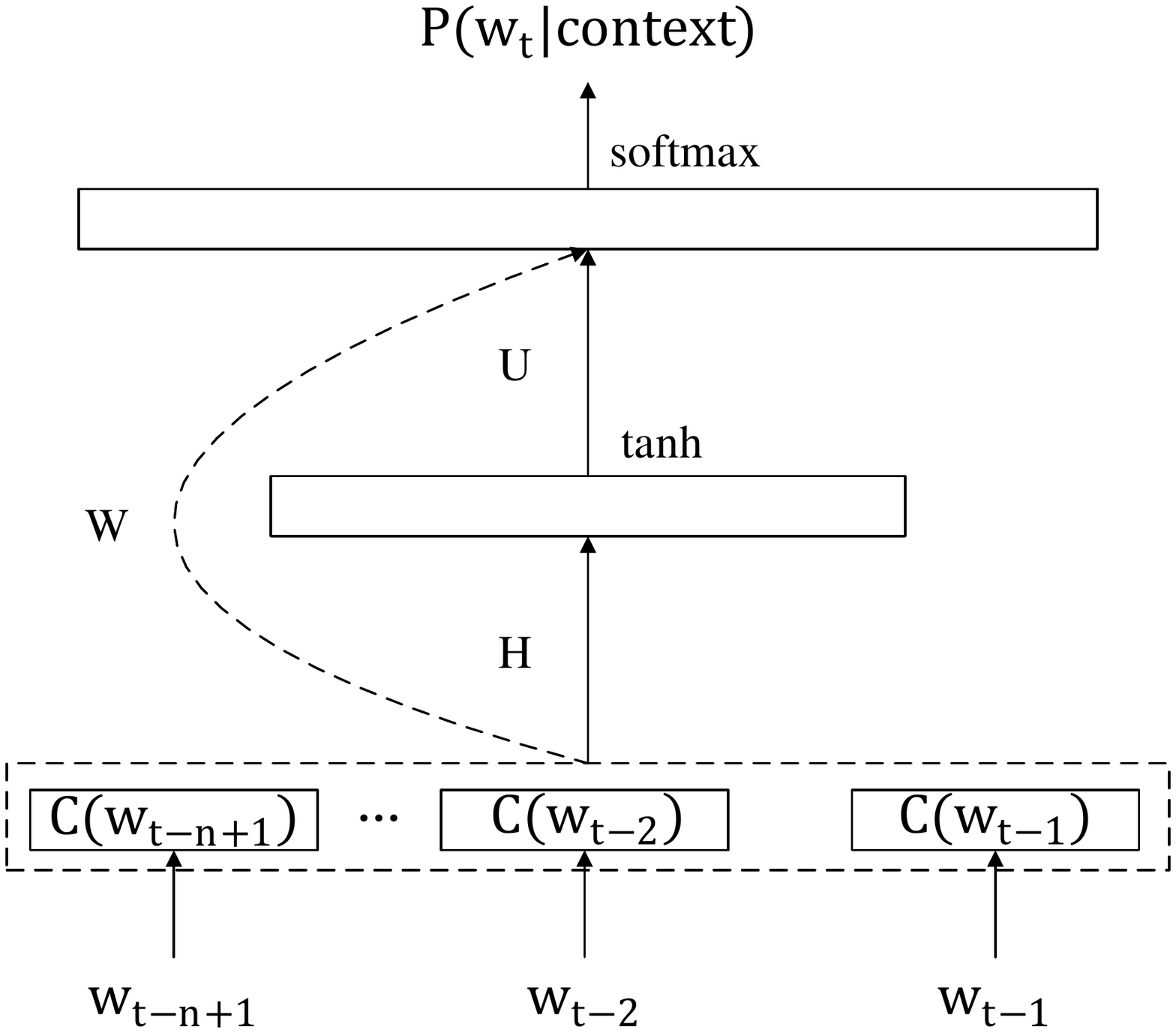}
	\caption{
        The FFNNLM proposed by [Bengio {\em et al}., 2003]. In this model, in order to predict the conditional probability of the word $w_t$, its previous $n-1$ words are projected by the shared projection matrix $C\in R^{|V|\times m}$ into a continuous feature vector space according to their index in the vocabulary, where $|V|$ is the size of the vocabulary and $m$ is the dimension of the feature vectors, i.e., the word $w_i$ is projected as the distributed feature vector $C(w_i)\in R^m$. Each row of the projection matrix $C$ is a feature vector of a word in the vocabulary. The input $x$ of the FFNN is a concatenation of feature vectors of $n-1$ words. This model is followed by Softmax output layer to guarantee all the conditional probabilities of words positive and summing to one. The learning algorithm is the Stochastic Gradient Descent (SGD) method using the backpropagation (BP) algorithm.}
    \label{fig:FFNNLM}
\end{figure}

\cite{bengio_neural_2003} proposed the architecture of the original FFNNLM, as shown in Figure \ref{fig:FFNNLM}. This FFNNLM can be expressed as:
\begin{align}
    y = b+Wx+Utanh(d+Hx),
\end{align}
where $H$, $U$, and $W$ are the weight matrixes that is for the connections between the layers; $d$ and $b$ are the biases of the hidden layer and the output layer.

FFNNLM implements modeling on continuous space by learning a distributed representation for each word. The word representation is a by-product of LMs, which is used to improve other NLP tasks. Based on FFNNLM, two word representation models, CBOW and Skip-gram , were proposed by \cite{mikolov_efficient_2013}. FFNNLM overcomes the curse of dimensions by converting words into low-dimensional vectors. FFNNLM leads the trend of NNLM research.

However, it still has a few drawbacks. The context size specified before training is limited, which is quite different from the fact that people can use lots of context information to make predictions. Words in a sequence are time-related. FFNNLM does not use timing information for modeling. Moreover, fully connected NN needs to learn many trainable parameters, even though these parameters are less than n-gram LM, which still is expensive and inefficient.
\subsection{RNN Language Models}
\cite{bengio_neural_2003} proposed the idea of using RNN for LMs. They claimed that introducing more structure and parameter sharing into NNs could capture longer contextual information.

\begin{figure}
	\centering
	\includegraphics[width=8cm,trim=0 170 0 170,scale=0.1]{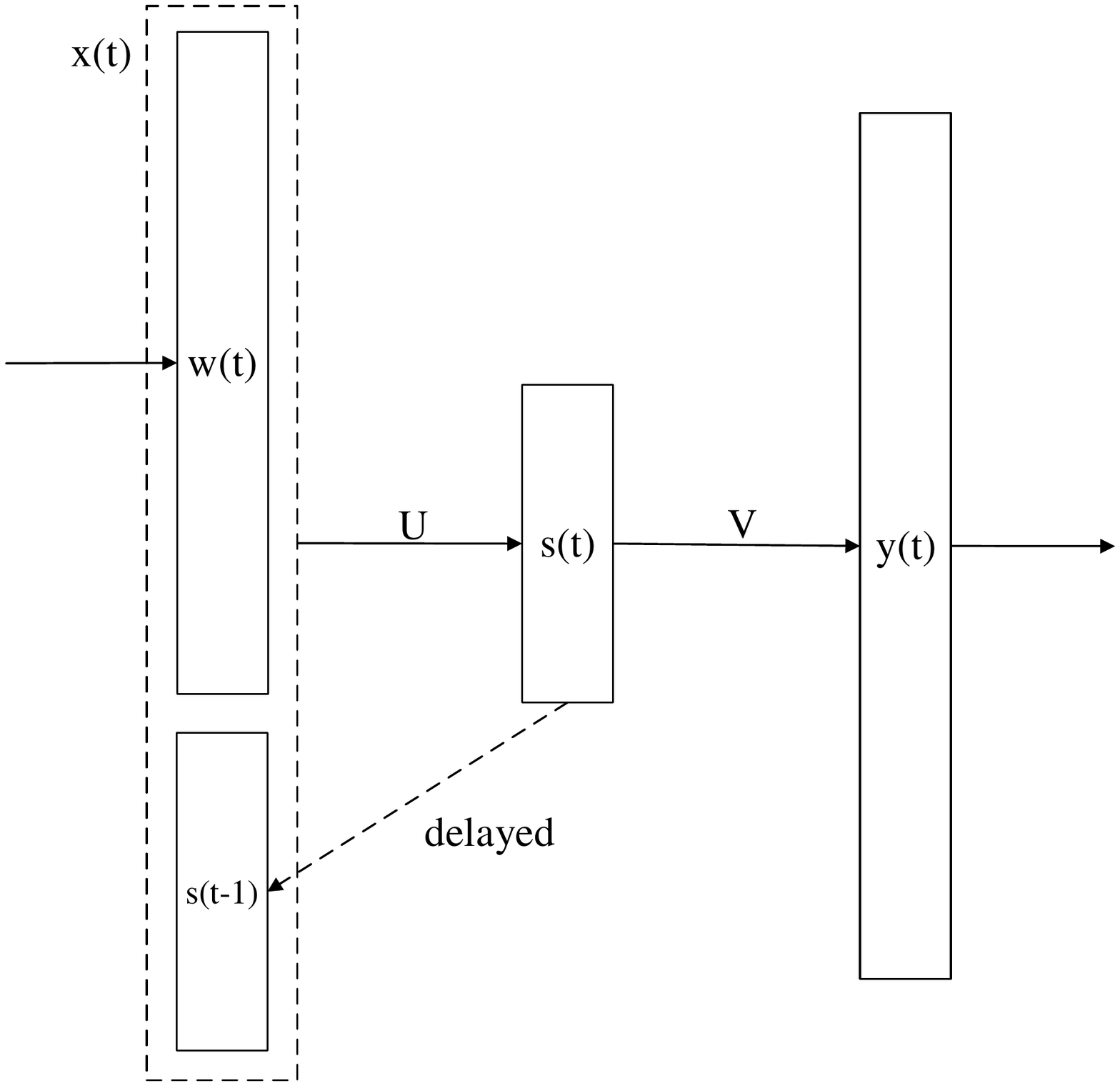}
	\caption{
        The RNNLM proposed by  [Mikolov {\em et al}., 2010; Mikolov {\em et al}., 2011]. The RNN has an internal state that changes with the input on each time step, taking into account all previous contexts. The state $s_t$ can be derived from the input word vector $w_t$ and the state $s_{t-1}$.}
    \label{fig:RNNLM}
\end{figure}

The first RNNLM was proposed  by \cite{mikolov_recurrent_2010,mikolov_extensions_2011}. As shown in Figure \ref{fig:RNNLM}, at time step $t$, the RNNLM can be described as:
\begin{align}
    x_t &= [w_t^\mathrm{T};s_{t-1}^\mathrm{T}]^\mathrm{T},\nonumber\\ s_t &= f(Ux_t+b),\nonumber\\ y_t &= g(Vs_t+d),
\end{align}
where $U$, $W$, $V$ are weight matrixes; $b$, $d$ are the biases of the state layer and the output layer respectively; in \cite{mikolov_recurrent_2010,mikolov_extensions_2011}, $f$ is the sigmoid function, and $g$ is the Softmax function. RNNLMs could be trained by the backpropagation through time (BPTT) or the truncated BPTT algorithm. According to their experiments, the RNNLM is significantly superior to FFNNLMs and n-gram LMs in terms of PPL.

Compared with FFNNLM, RNNLM has made a break-through. RNNs have an inherent advantage in the processing of sequence data as variable-length inputs could be received. When the network input window is shifted, duplicate calculations are avoided as the internal state. And the changes of the internal state by the input at $t$ step reveal timing information. Parameter sharing significantly reduces the number of parameters.

Although RNNLMs could take advantage of all contexts for prediction, it is challenging for training models to learn long-term dependencies. The reason is that the gradients of parameters can disappear or explode during RNN training, which leads to slow training or infinite parameter value.
\subsection{LSTM-RNN Language Models}
Long Short-Term Memory (LSTM) RNN solved this problem. \cite{sundermeyer_lstm_2012} introduced LSTM into LM and proposed LSTM-RNNLM. Except for the memory unit and the part of NN, the architecture of LSTM-RNNLM is almost the same as RNNLM. Three gate structures (including input, output, and forget gate) are added to the LSTM memory unit to control the flow of information. The general architecture of LSTM-RNNLM can be formalized as:
\begin{align}
    i_t &= \sigma(U_ix_t+W_is_{t-1}+V_ic_{t-1}+b_i),\nonumber\\
    f_t &= \sigma(U_fx_t+W_fs_{t-1}+V_fc_{t-1}+b_f),\nonumber\\
    g_t &= f(Ux_t+Ws_{t-1}+Vc_{t-1}+b),\nonumber\\
    c_t &= f_t\odot c_{t-1}+i_t\odot g_t,\nonumber\\
    o_t &= \sigma(U_ox_t+W_os_{t-1}+V_oc_t+b_o),\nonumber\\
    s_t &= o_t\cdot f(c_t),\nonumber\\
    y_t &= g(Vs_t+Mx_t+d),
\end{align}
where $i_t$, $f_t$, $o_t$ are input gate, forget gate and output gate, respectively. $c_t$ is the internal memory of unit. $s_t$ is the hidden state unit. $U_i$, $U_f$, $U_o$, $U$, $W_i$, $W_f$, $W_o$, $W$, $V_i$, $V_f$, $V_o$, and $V$ are all weight matrixes. $b_i$, $b_f$, $b_o$, $b$, and $d$ are bias. $f$ is the activation function and $\sigma$ is the activation function for gates, generally a sigmoid function.

Comparing the above three classic LMs, RNNLMs (including LSTM-RNNLM) perform better than FFNNLM, and LSTM-RNNLM maintains the state-of-the-art LM. The current NNLMs are based mostly on RNN or LSTM.

Although LSTM-RNNLM performs well, training the model on a large corpus is very time-consuming because the distributions of predicted words are explicitly normalized by the Softmax layer, which leads to considering all words in the vocabulary when computing the log-likelihood gradients. Also, better performance of LM is expected. To improve NNLM, researchers are still exploring different techniques that are similar to how humans process natural language.
\section{Improved Techniques}
\subsection{Techniques for Reducing Perplexity}
New structures and more effective information are introduced into classic NNLMs, especially LSTM-RNNLM, for reducing PPL. Inspired by linguistics and how humans process natural language, some novel, effective methods, including character-aware models, factored models, bidirectional models, caching, attention, etc., are proposed.
\subsubsection{Character-Aware Models}
In natural language, some words in the similar form often have the same or similar meaning. For example, {\em man} in {\em superman} has the same meaning as the one in {\em policeman}. \cite{mikolov_subword_2012} explored RNNLM and FFNNLM at the character level. Character-level NNLM can be used for solving out-of-vocabulary (OOV) word problem, improving the modeling of uncommon and unknown words because character features reveal structural similarities between words. Character-level NNLM also reduces training parameters due to the small Softmax layers with character-level outputs. However, the experimental results showed that it was challenging to train highly-accurate character-level NNLMs, and its performance is usually worse than the word-level NNLMs. This is because character-level NNLMs have to consider a longer history to predict the next word correctly.

A variety of solutions that combine character- and word-level information, generally called character-aware LM, have been proposed. One approach is to organize character-level features word by word and then use them for word-level LMs. \cite{kim_character-aware_2015} proposed Convolutional Neural Network (CNN) for extracting character-level feature and LSTM for receiving these character-level features of the word in a time step. \cite{hwang_character-level_2016} solved the problem of character-level NNLMs using a hierarchical RNN architecture consisting of multiple modules with different time scales. Another solution is to input character- and word-level features into NNLM simultaneously. \cite{miyamoto_gated_2016} suggested interpolating word feature vectors with character feature vectors extracted from words by BiLSTM and inputting interpolation vectors into LSTM. \cite{verwimp_character-word_2017} proposed a character-word LSTM-RNNLM that directly concatenated character- and word-level feature vectors and input concatenations into the network. Character-aware LM directly uses character-level LM as character feature extractor for word-level LM. Therefore, the LM has rich character-word information for prediction.
\subsubsection{Factored Models}
NNLMs define the similarity of words based on tokens. However, similarity can also be derived from word shape features (affixes, uppercase, hyphens, etc.) or other annotations (such as POS). Inspired by factored LMs, \cite{alexandrescu_factored_2006} proposed a factored NNLMs, a novel neural probabilistic LM that can learn the mapping from words and specific features of the words to continuous spaces.

Many studies have explored the selection of factors. Different linguistic features are considered first. \cite{wu_factored_2012} introduced morphological, grammatical, and semantic features to extend RNNLMs. \cite{adel_syntactic_2015} also explored the effects of other factors, such as part-of-speech tags, Brown word clusters, open class words, and clusters of open class word embeddings. Experimental results showed that Brown word clusters, part-of-speech tags, and open words are most effective for a Mandarin-English Code-Switching task. Contextual information also was explored. For example, \cite{mikolov_context_2012} used the distribution of topics calculated from fixed-length blocks of previous words. \cite{wang_larger-context_2015} proposed a new approach to incorporating corpus bag-of-words (BoW) context into language modeling. Besides, some methods based on text-independent factors were proposed. \cite{ahn_neural_2016} proposed a Neural Knowledge Language Model that applied the notation knowledge provided by knowledge graph for RNNLMs.

The factored model allows the model to summarize word classes with same characteristics. Applying factors other than word tokens to the neural network training can better learn the continuous representation of words, represent OOV words, and reduce the PPL of LMs. However, the selection of different factors is related to different upstream NLP tasks, applications, etc. of LM. And there is no way to select factors in addition to experimenting with various factors separately. Therefore, for a specific task, an efficient selection method of factors is necessary. Meanwhile, corpora with factor labels have to be established.
\subsubsection{Bidirectional Models}
Traditional unidirectional NNs only predict the outputs from past inputs. A bidirectional NN can be established, which is conditional on future data. \cite{graves_hybrid_2013,bahdanau_neural_2014} introduced bidirectional RNN and LSTM neural networks (BiRNN and BiLSTM) into speech recognition or other NLP tasks. The BiRNNs utilize past and future contexts by processing the input data in both directions. One of the most popular works of the bidirectional model is the ELMo model \cite{peters_deep_2018}, a new deep contextualized word representation based on BiLSTM-RNNLMs. The vectors of the embedding layer of a pre-trained ELMo model is the learned representation vector of words in the vocabulary. These representations are added as the embedding layer of the existing model and significantly improve state of the art across six challenging NLP tasks.

Although BiLM using past and future contexts has achieved improvements, it is noted that BiLM cannot be used directly for LM because LM is defined in the previous context. BiLMs can be used for other NLP tasks, such as machine translation, speech recognition because the word sequence is regarded as a simultaneous input sequence.
\subsubsection{Caching}
Words that have appeared recently may appear again. Due to this assumption, the cache mechanism initially is used to optimize n-gram LM, overcoming the length limit of dependencies. It matches a new input and histories in the cache. Cache mechanism was originally proposed for reducing PPL of NNLMs. \cite{hutchison_neural_2012} attempted to combine FFNNLM with the cache mechanism and proposed a structure of the cache-based NNLMs, which leads to discrete probability change. To solve this problem,  \cite{grave_improving_2016} proposed a continuous cache model, where the change depends on the inner product of the hidden representations.

Another type of cache mechanism is that cache is used as a speed-up technique for NNLMs. The main idea of this method is to store the outputs and states of LMs in a hash table for future prediction given the same contextual history. For example, \cite{huang_cache_2014} proposed the use of four caches to accelerate model reasoning. The caches are respectively Query to Language Model Probability Cache, History to Hidden State Vector Cache, History to Class Normalization Factor Cache, and History and Class Id to Sub-vocabulary Normalization Factor Cache.

The caching technique is proved that it can reduce computational complexity and improve the learning ability of long-term dependence due to its caches. It can be applied flexibly to existing models. However, if the size of the cache is limited, cache-based NNLM will not perform well.
\subsubsection{Attention}
RNNLMs predict the next word with its context. Not every word in the context is related to the next word and effective for prediction. Similar to human beings, LM with the attention mechanism uses the long history efficiently by selecting useful word representations from them. \cite{bahdanau_neural_2014} first proposed the application of the attention mechanism to NLP tasks (machine translation in this paper). \cite{tran_recurrent_2016,mei_coherent_2016} proved that the attention mechanism could improve the performance of RNNLMs.

The attention mechanism obtains the target areas that need to be focused on by a set of attention coefficients for each input. The attention vector $z_t$ is calculated by the representation $\{r_0,r_1,\cdots,r_{t-1}\}$ of tokens:
\begin{align}
    z_t=\sum_{i=0}^{t-1}\alpha_{ti}r_i.
\end{align}
This attention coefficient $\alpha_{ti}$ is normalized by the score $e_{ti}$ by the Softmax function, where
\begin{align}
    e_{ti}=a(h_{t-1},r_i),
\end{align}
is an alignment model that evaluates how well the representation $r_i$ of a token and the hidden state $h_{t-1}$ match. This attention vector is a good representation of the history of context for prediction.

The attention mechanism has been widely used in Computer Vision (CV) and NLP. Many improved attention mechanisms have been proposed, including soft/hard attention, global/local attention, key-value/key-value-predict attention, multi-dimensional attention, directed self-attention, self-attention, multi-headed attention, and so on. These improved methods are used in LMs and have improved the quality of LMs.

On the basis of the above improved attention mechanisms, some competitive LMs or methods of word representation are proposed. Transformer proposed by \cite{vaswani_attention_2017} is the basis for the development of the subsequent models. Transformer is a novel structure based entirely on the attention mechanism, which consists of an encoder and a decoder. Since then, GPT \cite{radford_improving_2018} and BERT \cite{devlin_bert:_2018} have been proposed. The main difference is that GPT uses Transformer's decoder, and BERT uses Transformer's encoder. BERT is an attention-based bidirectional model. Unlike CBOW, Skip-gram, and ELMo, GPT and BERT represent words through the parameters of the entire model. They are state-of-the-art methods of word representation in NLP.

Attention mechanism with its applications for various tasks is one of the most popular research directions. Although various structures of attention mechanism have been proposed, as the core of attention mechanism, the methods for calculating the similarity between words have still not been improved. Proposing some novel methods for vector similarity play an important role in improving attention mechanism.
\subsection{Speed-up Techniques on Large Corpora}
Training the model on a corpus with a large vocabulary is very time-consuming. The main reason is the Softmax layer for large vocabulary. Many approaches have been proposed to address the difficulty of training deep NNs with large output spaces. In general, they can be divided into four categories, i.e., hierarchical Softmax, sampling-based approximations, self-normalization, and exact gradient on limited loss functions. Among them, the former two are used widely in NNLMs.
\subsubsection{Hierarchical Softmax}
Some methods based on hierarchical Softmax that decompose target conditional probability into multiples of some conditional probabilities are proposed. \cite{morin_hierarchical_2005} used a hierarchical binary tree (by the similarity from Wordnet) of an output layer, in which the $V$ words in vocabulary are regarded as its leaves. This technique allows exponentially fast calculations of word probabilities and their gradients. However, it performs much worse than non-hierarchical one despite using expert knowledge. \cite{mnih_scalable_2008} improved it by a simple feature-based algorithm for automatically building word trees from data. The performance of the above two models is mostly dependent on the tree, which is usually heuristically constructed.

By relaxing the constraints of the binary tree structure, \cite{le_structured_2011} introduced a new, general, better class-based NNLM with a structured output layer, called Structured Output Layer NNLM. Given a history $h$ of a word $w_i$, the conditional probability can be formed as:
\begin{align}
    P(w_t|h)=P(c_1(w_t)|h)\prod_{d=2}^DP(c_d(w_t)|h,c_1,\cdots,c_{d-1}).
\end{align}
Since then, many scholars have improved this model. Hierarchical models based on word frequency classification \cite{mikolov_extensions_2011} and Brown clustering \cite{si_impact_2012} were proposed. It was proved that the model with Brown clustering performed better. \cite{zweig_speed_2013} proposed a speed optimal classification, i.e., a dynamic programming algorithm that determines the classes by minimizing the running time of the model.

Hierarchical Softmax significantly reduces model parameters without increasing PPL. The reason is that the technique uses $d+1$ Softmax layers with $\sqrt[d+1]{|V|}$ classes or $log_2|V|$ two classification instead of one Softmax layers with $|V|$ classes. Nevertheless, hierarchical Softmax leads NNLMs to perform worse. Hierarchical Softmax based on Brown clustering is a special case. At the same time, the existing class-based methods do not consider the context. The classification is hard classification, which is a key factor of the increase of PPL. Therefore, it is necessary to study a method based on soft classification. It is our hope that while reducing the cost of NNLM training, PPL will remain unchanged, even decrease.
\subsubsection{Sampling-based Approximations}
When the NNLMs calculate the conditional probability of the next word, the output layer uses the Softmax function, where the cost of calculation of the normalized denominator is extremely expensive. Therefore, one approach is to randomly or heuristically select a small portion of the output and estimate the probability and the gradient from the samples.

Inspired by Minimizing Contrastive Divergence, \cite{bengio_quick_2003} proposed an importance sampling method and an adaptive importance sampling algorithm to accelerate the training of NNLMs. The gradient of the log-likelihood can be expressed as:
\begin{align}
    \frac{\partial logP(w_t|w_1^{t-1})}{\partial\theta} = -& \frac{\partial y(w_t,w_1^{t-1})}{\partial\theta}\nonumber\\ +& \sum_{i=1}^kP(v_i|w_1^{t-1})\frac{\partial y(v_i,w_1^{t-1})}{\partial\theta}.
\end{align}
The weighted sum of the negative terms is obtained by importance sample estimates, i.e., sampling with $Q$ instead of $P$. Therefore, the estimates of the normalized denominator and the log-likelihood gradient are respectively:
\begin{align}
    \hat{Z}&(h_t) = \frac{1}{N}\sum_{w'~Q(\cdot|h_t)}\frac{e^{-y(w',h_t)}}{Q(w'|h_t)},\\
    E[&\frac{\partial logP(w_t|w_1^{t-1})}{\partial\theta}]= -\frac{\partial y(w_t,w_1^{t-1})}{\partial\theta}\nonumber\\ +& \frac{\sum_{w'\in\Gamma}\frac{\partial y(w',w_1^{t-1})}{\partial\theta}e^{-y(w',w_1^{t-1})}/Q(w'|w_1^{t-1})}{e^{-y(w',w_1^{t-1})}/Q(w'|w_1^{t-1})}.
\end{align}
Experimental results showed that adopting importance sampling leads to ten times faster the training of NNLMs without significantly increasing PPL. \cite{bengio_adaptive_2008} proposed an adaptive importance sampling method using an adaptive n-gram model instead of the simple unigram model in \cite{bengio_quick_2003}. Other improvements have been proposed, such as parallel training of small models to estimate loss for importance sampling, multiple importance sampling and likelihood weighting scheme, two-stage sampling, and so on. In addition, there are other different sampling methods for the training of NNLMs, including noise comparison estimation, Locality Sensitive Hashing (LSH) techniques, BlackOut.

These methods have significantly accelerated the training of NNLMs by sampling, while the model evaluation remains computationally challenging. At present, the computational complexity of the model with sampling-based approximations still is high. Sampling strategy is relatively simple. Except for LSH techniques, other strategies are to select at random or heuristically.
\section{Corpora}
As mentioned above, it is necessary to evaluate all LMs on the same corpus, but which is impractical. This section describes some of the common corpora.

In general, in order to reduce the cost of training and test, the feasibility of models needs to be verified on the small corpus first. Common small corpora include Brown, Penn Treebank, and WikiText-2 (see Table \ref{tab:smallcor}).

\begin{table}
\centering
\begin{tabular}{lrrrr}
\toprule
Corpus&Train &Valid  &Test &Vocab\\
\midrule
Brown         & 800,000  & 200,000 & 181,041 & 47,578  \\
Penn Treebank & 930,000  & 74,000  & 82,000  & 10,000  \\
WikiText-2    & 2000,000 & 220,000 & 250,000 & 33,000  \\
\bottomrule
\end{tabular}
\caption{Size (words) of small corpora}
\label{tab:smallcor}
\end{table}

After the model structure is determined, it needs to be trained and evaluated in a large corpus to prove that the model has a reliable generalization. Common large corpora that are updated over time from website, newspaper, etc. include Wall Street Journal, Wikipedia, News Commentary, News Crawl, Common Crawl, Associated Press (AP) News, and more.

However, LMs are often trained on different large corpora. Even on the same corpus, various preprocessing methods and different divisions of training/test set affect the results. At the same time, training time is reported in different ways or is not given in some papers. The experimental results in different papers do not be compared fully.
\section{Toolkits}
Traditional LM toolkits mainly includes CMU-Cambridge SLM, SRILM, IRSTLM, MITLM, BerkeleyLM, which only support the training and evaluation of n-gram LMs with a variety of smoothing. With the development of Deep Learning, many toolkits based on NNLMs are proposed. \cite{mikolov_rnnlm_2011} built the RNNLM toolkit, which supports the training of RNNLMs to optimize speech recognition and machine translation, but it does not support parallel training algorithms and GPU. \cite{schwenk_cslm_2013} constructed the neural network open source tool CSLM (Continuous Space Language Modeling) to support the training and evaluation of FFNNs. \cite{enarvi_theanolm_2016} proposed the scalable neural network model toolkit TheanoLM, which trains LMs to score sentences and generate text.

According to our survey, we found that there is no toolkit supporting both the traditional N-gram LM and NNLM. And they generally do not contain commonly used LM loads.
\section{Future Directions}
Most NNLMs are based on three classic NNLMs. LSTM-RNNLMs is the state-of-the-art LM. There are two directions of improving LMs, i.e., reducing PPL using a novel structure or an additional knowledge and accelerating the training and evaluation by estimate conditional probability.

During the survey, some existing problems in NNLMs are found and summarized. Therefore, we propose the future direction of LMs. Firstly, the methods that reduce the cost and the number of parameters would continue to be explored to speed up the training and evaluation without PPL increasing. Then, a novel structure to simulate the way humans work is expected for improving the performance of LM. For example, building a generative model, such as GAN, for LMs may be a new direction. Last but not least, the current evaluation system of LMs is not standardized. It is necessary to build an evaluation benchmark for unifying the preprocessing and what results should be reported in papers.
\section{Conclusion}
The study of NNLMs has been going on for nearly two decades. NNLMs have made timely and significant contributions to NLP tasks. The different architectures of the classic NNLMs and their improvement are surveyed. Their related corpora and toolkits that are essential for the study of NNLMs are also introduced.
\appendix
\bibliographystyle{named}
\bibliography{ijcai19}

\begin{thebibliography}{}

\bibitem[\protect\citeauthoryear{Adel \bgroup \em et al.\egroup
  }{2015}]{adel_syntactic_2015}
H.~Adel, N.~T. Vu, K.~Kirchhoff, D.~Telaar, and T.~Schultz.
\newblock Syntactic and semantic features for code-switching factored language
  models.
\newblock {\em {IEEE/ACM} Trans. Audio, Speech {\&} Language Processing},
  23(3):431--440, 2015.

\bibitem[\protect\citeauthoryear{Ahn \bgroup \em et al.\egroup
  }{2016}]{ahn_neural_2016}
S.~Ahn, H.~Choi, T.~P{\"{a}}rnamaa, and Y.~Bengio.
\newblock A neural knowledge language model.
\newblock {\em CoRR}, abs/1608.00318, 2016.

\bibitem[\protect\citeauthoryear{Alexandrescu and
  Kirchhoff}{2006}]{alexandrescu_factored_2006}
A.~Alexandrescu and K.~Kirchhoff.
\newblock Factored neural language models.
\newblock In {\em {HLT-NAACL}}, 2006.

\bibitem[\protect\citeauthoryear{Bahdanau \bgroup \em et al.\egroup
  }{2014}]{bahdanau_neural_2014}
D.~Bahdanau, K.~Cho, and Y.~Bengio.
\newblock Neural machine translation by jointly learning to align and
  translate.
\newblock {\em CoRR}, abs/1409.0473, 2014.

\bibitem[\protect\citeauthoryear{Bengio and Senecal}{2003}]{bengio_quick_2003}
Y.~Bengio and J.~Senecal.
\newblock Quick training of probabilistic neural nets by importance sampling.
\newblock In {\em {AISTATS}}, 2003.

\bibitem[\protect\citeauthoryear{Bengio and
  Senecal}{2008}]{bengio_adaptive_2008}
Y.~Bengio and J.~Senecal.
\newblock Adaptive importance sampling to accelerate training of a neural
  probabilistic language model.
\newblock {\em {IEEE} Trans. Neural Networks}, 19(4):713--722, 2008.

\bibitem[\protect\citeauthoryear{Bengio \bgroup \em et al.\egroup
  }{2003}]{bengio_neural_2003}
Y.~Bengio, R.~Ducharme, P.~Vincent, and C.~Janvin.
\newblock A neural probabilistic language model.
\newblock {\em {JMLR}}, 3:1137--1155, 2003.

\bibitem[\protect\citeauthoryear{Devlin \bgroup \em et al.\egroup
  }{2018}]{devlin_bert:_2018}
J.~Devlin, M.~Chang, K.~Lee, and K.~Toutanova.
\newblock {BERT:} pre-training of deep bidirectional transformers for language
  understanding.
\newblock {\em CoRR}, abs/1810.04805, 2018.

\bibitem[\protect\citeauthoryear{Enarvi and
  Kurimo}{2016}]{enarvi_theanolm_2016}
S.~Enarvi and M.~Kurimo.
\newblock Theanolm - an extensible toolkit for neural network language
  modeling.
\newblock In {\em Interspeech}, pages 3052--3056, 2016.

\bibitem[\protect\citeauthoryear{Grave \bgroup \em et al.\egroup
  }{2016}]{grave_improving_2016}
E.~Grave, A.~Joulin, and N.~Usunier.
\newblock Improving neural language models with a continuous cache.
\newblock {\em CoRR}, abs/1612.04426, 2016.

\bibitem[\protect\citeauthoryear{Graves \bgroup \em et al.\egroup
  }{2013}]{graves_hybrid_2013}
A.~Graves, N.~Jaitly, and A.~Mohamed.
\newblock Hybrid speech recognition with deep bidirectional {LSTM}.
\newblock In {\em {IEEE} Workshop on ASRU}, pages 273--278, 2013.

\bibitem[\protect\citeauthoryear{Huang \bgroup \em et al.\egroup
  }{2014}]{huang_cache_2014}
Z.~Huang, G.~Zweig, and B.~Dumoulin.
\newblock Cache based recurrent neural network language model inference for
  first pass speech recognition.
\newblock In {\em {IEEE} {ICASSP}}, pages 6354--6358, 2014.

\bibitem[\protect\citeauthoryear{Hwang and
  Sung}{2016}]{hwang_character-level_2016}
K.~Hwang and W.~Sung.
\newblock Character-level language modeling with hierarchical recurrent neural
  networks.
\newblock {\em CoRR}, abs/1609.03777, 2016.

\bibitem[\protect\citeauthoryear{Jelinek \bgroup \em et al.\egroup
  }{1977}]{jelinek_perplexitymeasure_1977}
F.~Jelinek, R.~L. Mercer, L.~R. Bahl, and J.~K. Baker.
\newblock Perplexity—a measure of the difficulty of speech recognition tasks.
\newblock {\em {JASA}}, 62(S1):S63--S63, December 1977.

\bibitem[\protect\citeauthoryear{Kim \bgroup \em et al.\egroup
  }{2015}]{kim_character-aware_2015}
Y.~Kim, Y.~Jernite, D.~Sontag, and A.~M. Rush.
\newblock Character-aware neural language models.
\newblock {\em CoRR}, abs/1508.06615, 2015.

\bibitem[\protect\citeauthoryear{Le \bgroup \em et al.\egroup
  }{2013}]{le_structured_2011}
H.~S. Le, I.~Oparin, A.~Allauzen, J.~Gauvain, and F.~Yvon.
\newblock Structured output layer neural network language models for speech
  recognition.
\newblock {\em {IEEE} Trans. Audio, Speech {\&} Language Processing},
  21(1):195--204, 2013.

\bibitem[\protect\citeauthoryear{Mei \bgroup \em et al.\egroup
  }{2016}]{mei_coherent_2016}
H.~Mei, M.~Bansal, and M.~R. Walter.
\newblock Coherent dialogue with attention-based language models.
\newblock {\em CoRR}, abs/1611.06997, 2016.

\bibitem[\protect\citeauthoryear{Mikolov and
  Zweig}{2012}]{mikolov_context_2012}
T.~Mikolov and G.~Zweig.
\newblock Context dependent recurrent neural network language model.
\newblock In {\em {IEEE} {SLT}}, pages 234--239, 2012.

\bibitem[\protect\citeauthoryear{Mikolov \bgroup \em et al.\egroup
  }{2010}]{mikolov_recurrent_2010}
T.~Mikolov, M.~Karafi{\'{a}}t, L.~Burget, J.~Cernock{\'{y}}, and S.~Khudanpur.
\newblock Recurrent neural network based language model.
\newblock In {\em {INTERSPEECH}}, pages 1045--1048, 2010.

\bibitem[\protect\citeauthoryear{Mikolov \bgroup \em et al.\egroup
  }{2011a}]{mikolov_extensions_2011}
T.~Mikolov, S.~Kombrink, L.~Burget, J.~Cernock{\'{y}}, and S.~Khudanpur.
\newblock Extensions of recurrent neural network language model.
\newblock In {\em Proc. of {IEEE} {ICASSP}}, pages 5528--5531, 2011.

\bibitem[\protect\citeauthoryear{Mikolov \bgroup \em et al.\egroup
  }{2011b}]{mikolov_rnnlm_2011}
T.~Mikolov, S.~Kombrink, A.~Deoras, and L.~Burget.
\newblock {RNNLM} - {Recurrent} {Neural} {Network} {Language} {Modeling}
  {Toolkit}.
\newblock In {\em {IEEE} {ASRU}}, page~4, 2011.

\bibitem[\protect\citeauthoryear{Mikolov \bgroup \em et al.\egroup
  }{2012}]{mikolov_subword_2012}
T.~Mikolov, I.~Sutskever, A.~Deoras, H.~Le, and S.~Kombrink.
\newblock {Subword} {Language} {Modeling} {With} {Neural} {Networks}.
\newblock 2012.

\bibitem[\protect\citeauthoryear{Mikolov \bgroup \em et al.\egroup
  }{2013}]{mikolov_efficient_2013}
T.~Mikolov, K.~Chen, G.~Corrado, and J.~Dean.
\newblock Efficient estimation of word representations in vector space.
\newblock {\em CoRR}, abs/1301.3781, 2013.

\bibitem[\protect\citeauthoryear{Miyamoto and Cho}{2016}]{miyamoto_gated_2016}
Y.~Miyamoto and Ky. Cho.
\newblock Gated word-character recurrent language model.
\newblock In {\em Proc. of {EMNLP}}, pages 1992--1997, 2016.

\bibitem[\protect\citeauthoryear{Mnih and Hinton}{2008}]{mnih_scalable_2008}
A.~Mnih and G.~E. Hinton.
\newblock A scalable hierarchical distributed language model.
\newblock In {\em Proc. of NIPS}, pages 1081--1088, 2008.

\bibitem[\protect\citeauthoryear{Morin and
  Bengio}{2005}]{morin_hierarchical_2005}
F.~Morin and Y.~Bengio.
\newblock Hierarchical probabilistic neural network language model.
\newblock In {\em Proc. of {AISTATS}}, 2005.

\bibitem[\protect\citeauthoryear{Peters \bgroup \em et al.\egroup
  }{2018}]{peters_deep_2018}
M.~E. Peters, M.~Neumann, M.~Iyyer, M.~Gardner, C.~Clark, K.~Lee, and
  L.~Zettlemoyer.
\newblock Deep contextualized word representations.
\newblock In {\em Proc. of {NAACL-HLT} Volume 1 (Long Papers)}, pages
  2227--2237, 2018.

\bibitem[\protect\citeauthoryear{Radford \bgroup \em et al.\egroup
  }{2018}]{radford_improving_2018}
A.~Radford, K.~Narasimhan, T.~Salimans, and I.~Sutskever.
\newblock Improving {Language} {Understanding} by {Generative}
  {Pre}-{Training}.
\newblock 2018.

\bibitem[\protect\citeauthoryear{Schwenk}{2013}]{schwenk_cslm_2013}
H.~Schwenk.
\newblock {CSLM} - a modular open-source continuous space language modeling
  toolkit.
\newblock In {\em {INTERSPEECH}, {ISCA}}, pages 1198--1202, 2013.

\bibitem[\protect\citeauthoryear{Si \bgroup \em et al.\egroup
  }{2012}]{si_impact_2012}
Y.~Si, Y.~Guo, Y.~Liu, J.~Pan, and Y.~Yan.
\newblock Impact of {Word} {Classing} on {Recurrent} {Neural} {Network}
  {Language} {Model}.
\newblock In {\em {GCIS}}, pages 100--103, November 2012.

\bibitem[\protect\citeauthoryear{Soutner \bgroup \em et al.\egroup
  }{2012}]{hutchison_neural_2012}
D.~Soutner, Z.~Loose, L.~M{\"{u}}ller, and A.~Praz{\'{a}}k.
\newblock Neural network language model with cache.
\newblock In {\em {TSD}}, pages 528--534. 2012.

\bibitem[\protect\citeauthoryear{Sundermeyer \bgroup \em et al.\egroup
  }{2012}]{sundermeyer_lstm_2012}
M.~Sundermeyer, R.~Schl{\"{u}}ter, and H.~Ney.
\newblock {LSTM} neural networks for language modeling.
\newblock In {\em {INTERSPEECH}}, pages 194--197, 2012.

\bibitem[\protect\citeauthoryear{Tran \bgroup \em et al.\egroup
  }{2016}]{tran_recurrent_2016}
K.~M. Tran, A.~Bisazza, and C.~Monz.
\newblock Recurrent memory networks for language modeling.
\newblock In {\em {NAACL} {HLT}}, pages 321--331, 2016.

\bibitem[\protect\citeauthoryear{Vaswani \bgroup \em et al.\egroup
  }{2017}]{vaswani_attention_2017}
A.~Vaswani, N.~Shazeer, N.~Parmar, J.~Uszkoreit, L.~Jones, A.~N. Gomez,
  L.~Kaiser, and I.~Polosukhin.
\newblock Attention is all you need.
\newblock In {\em {NIPS}}, pages 6000--6010, 2017.

\bibitem[\protect\citeauthoryear{Verwimp \bgroup \em et al.\egroup
  }{2017}]{verwimp_character-word_2017}
L.~Verwimp, J.~Pelemans, H.~Van hamme, and P.~Wambacq.
\newblock Character-word {LSTM} language models.
\newblock In {\em Proc. of {EACL} Volume 1: Long Papers}, pages 417--427, 2017.

\bibitem[\protect\citeauthoryear{Wang and Cho}{2015}]{wang_larger-context_2015}
T.~Wang and K.~Cho.
\newblock Larger-context language modelling.
\newblock {\em CoRR}, abs/1511.03729, 2015.

\bibitem[\protect\citeauthoryear{Wu \bgroup \em et al.\egroup
  }{2012}]{wu_factored_2012}
Y.~Wu, X.~Lu, H.~Yamamoto, S.~Matsuda, C.~Hori, and H.~Kashioka.
\newblock Factored language model based on recurrent neural network.
\newblock In {\em {COLING}, Proc. of the Conference: Technical Papers}, pages
  2835--2850, 2012.

\bibitem[\protect\citeauthoryear{Xu and Rudnicky}{2000}]{xu_can_2000}
W.~Xu and A.~Rudnicky.
\newblock Can artificial neural networks learn language models?
\newblock In {\em {ICSLP} / {INTERSPEECH}}, pages 202--205, 2000.

\bibitem[\protect\citeauthoryear{Zweig and Makarychev}{2013}]{zweig_speed_2013}
G.~Zweig and K.~Makarychev.
\newblock Speed regularization and optimality in word classing.
\newblock In {\em {IEEE} {ICASSP}}, pages 8237--8241, 2013.

\end{thebibliography}
\end{document}